\documentclass[letterpaper, 10pt, conference]{ieeeconf}
\IEEEoverridecommandlockouts
\usepackage{amsmath}
\usepackage{amsfonts}
\usepackage{amssymb}
\usepackage{graphicx}
\usepackage{siunitx}
\usepackage{booktabs}
\usepackage[table]{xcolor}
\usepackage{multirow}
\usepackage{algorithm}
\usepackage{algpseudocode}
\usepackage{cite}
\usepackage{url}
\usepackage{hyperref}
\hypersetup{hidelinks}

\makeatletter
\renewcommand\fs@ruled{\def\@fs@cfont{\bfseries}\let\@fs@capt\floatc@ruled
  \def\@fs@pre{\kern10pt\hrule height.8pt depth0pt \kern2pt}%
  \def\@fs@post{\kern2pt\hrule\relax}%
  \def\@fs@mid{\kern2pt\hrule\kern2pt}%
  \let\@fs@iftopcapt\iftrue}
\makeatother
\floatstyle{ruled}
\restylefloat{algorithm}

\title{\LARGE \bf Watch Your Step: Anticipated Perception for Humanoid Locomotion}

\author{Matthias Heyrman$^{1,*}$, Chenhao Li$^{1,2}$, Ioannis Dadiotis$^{1}$, Marco Hutter$^{1}$%
\thanks{This work has been submitted to the IEEE for possible publication. Copyright may be transferred without notice, after which this version may no longer be accessible.}%
\thanks{$^{1}$Robotic Systems Laboratory, ETH Z\"urich, Switzerland.}%
\thanks{$^{2}$ETH AI Center, ETH Z\"urich, Switzerland.}%
\thanks{$^{*}$Corresponding author: {\tt\footnotesize mheyrman@ethz.ch}}%
\thanks{Project page: \protect\href{https://leggedrobotics.github.io/watch-your-step}{\protect\nolinkurl{leggedrobotics.github.io/watch-your-step}}}%
}

\makeatletter
\long\def\@makecaption#1#2{%
\ifx\@captype\@IEEEtablestring%
\begin{center}{\footnotesize #1}\\{\footnotesize #2}\end{center}%
\@IEEEtablecaptionsepspace%
\else
\@IEEEfigurecaptionsepspace%
\setbox\@tempboxa\hbox{\footnotesize #1.~~ #2}%
\ifdim \wd\@tempboxa >\hsize%
\setbox\@tempboxa\hbox{\footnotesize #1.~~ }%
\parbox[t]{\hsize}{\footnotesize \noindent\unhbox\@tempboxa#2}%
\else%
\ifcenterfigcaptions \hbox to\hsize{\footnotesize\hfil\box\@tempboxa\hfil}%
\else \hbox to\hsize{\footnotesize\box\@tempboxa\hfil}%
\fi\fi\fi}
\makeatother

\begin{document}

\bstctlcite{IEEEexample:BSTcontrol}

\maketitle
\thispagestyle{empty}
\pagestyle{empty}

\begin{abstract}
Perceptive locomotion policies typically observe a dense robot-centric elevation map, although only the terrain beneath and around the next footsteps informs the control decision.
Learned encoders compress the whole visual input, leaving the policy to work out which inputs matter while learning the baseline locomotion task.
Learning both at once slows training and, without computationally expensive attention mechanisms, makes generalization to new terrain harder.
We instead remove redundant inputs from the elevation map before the policy sees it.
A lightweight foothold prediction model (FPM) predicts future footholds from proprioception and the velocity command alone, and we sample the elevation map only in small patches around these predicted footholds.
We train the prediction model alongside the locomotion policy, periodically retraining it on new policy rollouts as the robot's gait matures.
This allows the perception to anticipate where the policy will step, while the policy learns to leverage terrain information around those steps.
With the proposed lightweight preprocessing, the policy reaches a given terrain difficulty in roughly a third of the training iterations required with the complete map.
It outperforms policies that observe the complete map where footholds are sparse, including in narrow walled corridors, and on unseen terrain, and it matches an attention-based encoder trained at several times the cost.
The policy transfers zero-shot to a Unitree G1 humanoid traversing various obstacles, stairs, gaps, and step-downs beyond the training distribution in both indoor and outdoor environments.
\end{abstract}

\section{Introduction}

Vision is an indispensable tool when traversing unstructured terrain.
People constantly survey the ground ahead of them to plan their next steps and to adjust the trajectory of a foot in mid-air.
Deep reinforcement learning (RL) has driven significant progress in legged robot locomotion~\cite{pmlr-v164-rudin22a, doi:10.1126/scirobotics.abk2822}.
Using high-fidelity physics simulators, modern RL algorithms with parallelized environments, and effective techniques to reduce the sim-to-real gap, humanoids can be conditioned on a robot-centric elevation map~\cite{elevation_mapping} to traverse rough terrain.

However, this reliance on dense environmental maps introduces fundamental limitations related to sample efficiency, redundancy, and distraction.
Only the small regions of terrain where the feet will next make contact should inform control decisions, yet policies typically sample and process the full map uniformly, even though most of it describes terrain the robot will never step on, such as cells far to the side or directly opposite the commanded velocity direction.
These irrelevant inputs make it hard to learn which regions of the map matter and can distract the policy with terrain features that have no bearing on the next step, while the larger state space makes generalization to different terrains harder and increases the computational cost of the policy.
By contrast, people tend to look at the ground approximately 1.5 seconds, i.e., 2 to 3 steps, ahead of them, focusing on their planned footholds instead of the entire surrounding scene~\cite{FootPlacementLookControl}.
On more challenging terrains, this look-ahead lets them intuitively adapt a step in mid-air: if a foothold seems insecure or they notice a better spot near the original target, they shift their focus and change the step trajectory.

\begin{figure}[t]
    \centering
    \includegraphics[width=\linewidth]{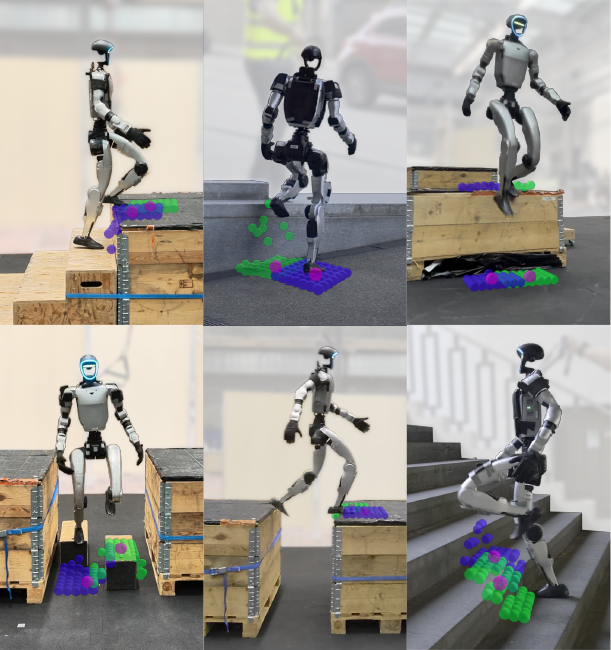}
    \vspace{-15pt}
    \caption{Zero-shot multi-terrain locomotion, including stairs, drops, gaps, unstable terrain, and crowded elevation-mapped environments using a Foothold Prediction Model (FPM) filtered elevation map. Predicted future footholds are shown in magenta, and the sampled elevation map points are shown in green and blue.}
    \label{fig:compiled_demo}
\end{figure}

Existing work primarily aims to reduce redundant information from the full elevation map through learned embeddings that feed the control policy.
Typical encoders compress the full input into a latent vector~\cite{doi:10.1126/scirobotics.abk2822}, scale well to complex terrains~\cite{hoeller2024anymalparkour, cheng2023parkour}, and transfer to real-world locomotion~\cite{long2024perceptiveinternalmodel, ben2025gallant}.
However, encoders such as CNNs still have to learn to extract the relevant features from the full map, and thus inherit the redundancy and sample inefficiency of the visual input rather than addressing it.

Attention-based encoders, conditioned on proprioception and the elevation map, weigh terrain features to focus on possible future steps~\cite{doi:10.1126/scirobotics.adv3604}, and later work extends them with local and global feature streams~\cite{zhang2026ame2agilegeneralizedlegged, fu2026globallocalattentiondecompositionterrain} and uncertainty-aware terrain estimates~\cite{zhang2026ame2agilegeneralizedlegged}.
Using egocentric depth images alongside a height scan enables an attention-based anticipatory gaze that better respects onboard compute limitations~\cite{li2026taga}.
These methods learn where to look, but nothing in the encoder structure ensures that the attended features are future footholds; the encoder has to discover the focus implicitly from the task reward.
Other depth-based methods encode the images directly~\cite{zhuang2024humanoid, yu2026ssr}.
All of these methods process the entire visual input when deployed and leave it to the policy to ignore what does not matter.
Terrain outside of its next steps can therefore still distract it.

Another approach analyzes the terrain and plans where the robot's feet should land, a long-standing body of work.
Classical methods select candidate contact regions based on kinematic reachability and terrain geometry~\cite{6907083, KANOULAS201913, 10.1109/TRO.2022.3222958, WANG2024104742}, and learning-based approaches evaluate candidate footholds based on the height map around the nominal landing position~\cite{magana2019footholdcnn} or across the local map under kinematic and collision constraints~\cite{belter2019singleshot}.
These foothold selection methods run as standalone modules that hand a target to a downstream tracking controller.
This requires strongly defined hand-crafted selection criteria for each platform, such as thresholds on the slope, roughness and curvature of a foot-sized patch, its distance to edges, and robot-specific checks of kinematic reachability and leg collision.
These criteria encode the designer's notion of a safe foothold and must be manually retuned for a new foot geometry or gait.
The module is also not necessarily conditioned on how the downstream controller actually moves, which leaves its output liable to be inaccurate or suboptimal for the gait that executes it.

The mismatch between planner and controller is especially pertinent for modern legged locomotion controllers, which are RL policies trained in simulation to maximize a reward.
RL-based approaches can learn precise foot placement in simulation through a terrain difficulty curriculum and crafted rewards~\cite{xie2020allsteps, wang2025beamdojo} and can learn to track foothold planners~\cite{jenelten2024dtc, montenegro2026mindstepsgenerallearning}, use them to shape rewards~\cite{dai2026walkplancphysicsguidedrl}, or track a decoupled learned planning policy~\cite{2019deepgait}.
In these cases, however, the foothold serves as a tracking target or a reward that shapes the policy's actions, not as a filter that shapes the policy's observations.
Prior work has used predicted footholds to reward safe locomotion using privileged information acquired in simulation during training~\cite{yu2026ssr}, but the predictions never reach the hardware.

Filtering the elevation map around future footholds before the policy sees it raises two challenges.
First, the filter cannot observe the terrain to decide where to look, as it would inherit the overhead of the full elevation map that it is meant to remove.
It must therefore predict the footholds entirely from the robot's own state and command, while the policy, like a person, can still adjust its trajectory to choose footholds based on the terrain it sees around each prediction.
Second, where the feet will land is a property of the gait, which the policy is still learning.
A predictor fit to an early, unsteady gait would misplace its patches once the policy improves, and a policy trained on misplaced patches may learn to distrust them.

In this paper, we propose a human-inspired, state-based sampling module that filters the elevation map before the locomotion policy receives any terrain information.
To filter the elevation map without observing it, a lightweight \emph{foothold prediction model} (FPM) predicts the robot's next two footholds from its current proprioceptive state and velocity command alone.
We treat only the terrain around these predicted footholds as relevant for locomotion and sample the elevation map in patches surrounding them.
These patches act as the exteroceptive observation of the locomotion policy, creating a state-conditioned representation of the terrain.

To ensure that the predicted footholds, and with them the sampled patches, stay up-to-date as the locomotion policy learns more difficult terrains, we co-train the FPM and the locomotion policy in an iterative training loop.
We train the locomotion policy with RL in simulation and, after a fixed number of iterations, refit the FPM in a self-supervised fashion on the footholds the current policy produces, allowing the perceptual filter to adapt as the policy converges.
We demonstrate that this approach reaches a given terrain difficulty in roughly a third of the iterations required by other methods using the complete elevation map, at a lower perception cost to the policy and across a variety of rough and sparse terrain styles.
We also show indoor and outdoor hardware experiments, where the resulting policy successfully transfers to a Unitree G1 humanoid (Fig.~\ref{fig:compiled_demo}).

In summary, our contributions are as follows:
\begin{enumerate}
    \item A self-supervised foothold prediction model (FPM) that accurately predicts the next two footholds from the current proprioceptive state and velocity command, co-trained on policy rollouts during RL training.
    \item A human-inspired perceptive locomotion pipeline that samples the elevation map only around the footholds predicted by the FPM, with insights into its key design choices, such as the prediction horizon and the encoding of the sampled patches.
    \item Simulation and hardware experiments, including comparisons against state-of-the-art elevation-map encoders, that demonstrate the advantages of restricting the policy's exteroceptive input to the terrain around its next predicted footholds over the complete map.
\end{enumerate}

\section{Preliminaries}
\label{sec:prelim}

We formulate the locomotion problem as an MDP $\mathcal{M} = \left(\mathcal{S}, \mathcal{A}, P, R, \gamma \right)$ where $\mathcal{S}$ denotes the robot's state space, $\mathcal{A}$ the robot's possible action space, $P(\mathbf{s}_{t+1}|\mathbf{s}_t, \mathbf{a}_t)$ represents the environment transition dynamics, $R(\mathbf{s}_t, \mathbf{a}_t)$ is the reward function, and $\gamma$ the discount factor.
We train the policy with PPO~\cite{PPO} across $8192$ parallel environments in IsaacLab~\cite{nvidia2025isaaclabgpuacceleratedsimulation} on a single NVIDIA RTX 3090 GPU, controlling a $29$-DoF Unitree G1 humanoid.
Physics is integrated at \SI{200}{\hertz} and the policy control rate is \SI{50}{\hertz}.

We define the system state $\mathbf{x}_t$ as the robot's estimated proprioceptive state:
\begin{equation}
    \mathbf{x}_t = \left[ \mathbf{v}_t,\ \boldsymbol{\omega}_t,\ \mathbf{g}_t,\ \mathbf{v}_{\mathrm{cmd}},\ \mathbf{q}_t,\ \dot{\mathbf{q}}_t \right] \in \mathbb{R}^{70},
    \label{eq:sysstate}
\end{equation}
composed of the base linear and angular velocities $\mathbf{v}_t$ and $\boldsymbol{\omega}_t$, respectively, the projected gravity vector $\mathbf{g}_t$, the velocity command $\mathbf{v}_{\mathrm{cmd}}$, and the joint positions and velocities $\mathbf{q}_t$ and $\dot{\mathbf{q}}_t$ relative to the nominal joint configuration.
The velocity command $\mathbf{v}_{\mathrm{cmd}} = (v_x, v_y, \omega_{\mathrm{yaw}}) \in [-1, 1]$\,\si{\metre\per\second}\,$\times [-0.5, 0.5]$\,\si{\metre\per\second}\,$\times [-1, 1]$\,\si{\radian\per\second} is resampled during training, with $2\%$ of the environments receiving a turn-in-place command and another $2\%$ a zero command.
We do not include the exteroceptive input in the robot state, treating it as a separate observation.

\begin{figure*}[t]
    \vspace{5pt}
    \centering
    \includegraphics[width=\linewidth]{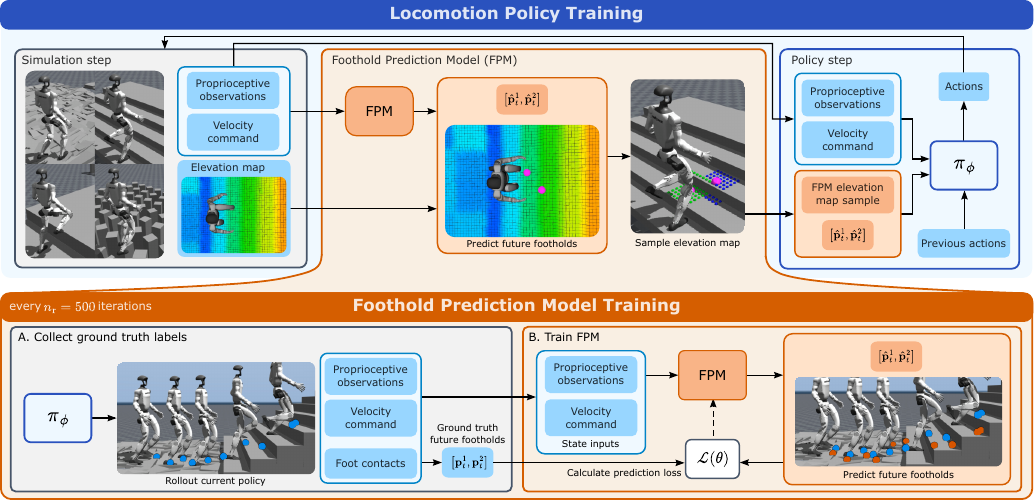}
    \vspace{-15pt}
    \caption{Method overview. The policy learns to track velocity commands over varied terrain from an elevation map gated around the FPM's predicted footholds. Every $n_\mathrm{r}=500$ iterations, A. rollouts of the current policy yield ground-truth future footholds and B. the FPM is retrained from scratch to predict them.}
    \label{fig:method}
\end{figure*}

\begin{algorithm}[t]
\footnotesize
\caption{Policy and FPM co-training}
\label{alg:training}
\begin{algorithmic}[1]

\State Initialize locomotion policy $\pi_\phi$ and FPM $f_\theta$
\State Pin $\hat{\mathbf{p}}_t^{\mathrm{1}}, \hat{\mathbf{p}}_t^{\mathrm{2}}$ to $(0, \pm 0.1)\,\si{\metre}$ beneath the base for bootstrapping
\State Train $\pi_\phi$ for $n_0 = 500$ iterations

\While{training}
    \State $K \gets \lceil N_{\mathrm{s}} / N_{\mathrm{env}} \rceil$ \Comment{$N_{\mathrm{s}} = 5\!\times\!10^5$ samples}
    \For{$K$ steps}
        \State Record $\{\mathbf{x}_t, c_t^{\mathrm{L}}, c_t^{\mathrm{R}}, \mathbf{r}_t\}$ using policy $\pi_\phi$
    \EndFor
    \State Label strictly future footholds $\mathbf{p}_t^{\mathrm{1}}, \mathbf{p}_t^{\mathrm{2}}$ to form $\mathcal{D}$
    \State Mask windows that span a termination (base-pose jump $> \SI{0.5}{\metre}$)
    \State Retrain $\theta$ from scratch on $\mathcal{D}$, $25$ epochs, $80/20$ split
    \State Export $f_\theta$; hot-swap into the observation pipeline

    \For{$n_{\mathrm{r}} = 500$ policy iterations}
        \State Predict footholds $\hat{\mathbf{p}}_t = f_\theta(\mathbf{x}_t)$
        \State Sample $k \times k$ height patches at $\hat{\mathbf{p}}_t^{\mathrm{1}}, \hat{\mathbf{p}}_t^{\mathrm{2}}$
        \State Update $\pi_\phi$ with PPO
        \State Advance terrain and assistive-force curricula
    \EndFor
\EndWhile
\end{algorithmic}
\end{algorithm}

\section{Method}
\label{sec:method}

Our framework, shown in Fig.~\ref{fig:method}, consists of two learned components: the FPM $f_\theta$, with parameters $\theta$, that predicts the next two footholds from the current system state $\mathbf{x}_t$, and a locomotion policy $\pi_\phi$, with parameters $\phi$, whose exteroceptive input is the elevation map sampled around only those predicted footholds.
We co-train the two components as summarized in Alg.~\ref{alg:training}, using PPO to continuously optimize the locomotion policy and refitting, in a self-supervised fashion, the FPM at fixed intervals to rollouts of the current policy.

\subsection{Foothold Prediction Model (FPM)}
\label{sec:fp}

The FPM models the closed-loop gait of the current locomotion policy,
\begin{equation*}
    f_{\theta}: \mathbb{R}^{70} \to \mathbb{R}^{4}, \qquad
    \hat{\mathbf{p}}_t = \left[\hat{\mathbf{p}}_t^{\mathrm{1}}, \hat{\mathbf{p}}_t^{\mathrm{2}}\right] = f_\theta(\mathbf{x}_t),
\end{equation*}
mapping the unscaled, noise-free system state $\mathbf{x}_t$ of Eq.~\eqref{eq:sysstate} to the planar positions $\hat{\mathbf{p}}_t^{\mathrm{1}}, \hat{\mathbf{p}}_t^{\mathrm{2}} \in \mathbb{R}^2$ of the next two footholds in the current base frame.
It is deliberately blind: it observes neither the elevation map nor its own previous output.

\subsubsection{Prediction target}
\label{ssec:fp_target}
As the robot moves during the policy rollouts of the co-training loop, we record its state trajectory together with every new foothold after a foot is airborne, detected from the binary contact indicators $c_t^{\mathrm{L}}, c_t^{\mathrm{R}}$.
Once the rollout is complete, we look backward in time and pair each past state $\mathbf{x}_t$ with the planar positions of the next two footholds in this sequence, irrespective of which foot made them, expressed in the base frame at time $t$ as the label $\mathbf{p}_t = [\mathbf{p}_t^\mathrm{1}, \mathbf{p}_t^\mathrm{2}]$.
The first foothold is the landing location of the foot in the air or about to lift off, and the second is the first contact of the other foot \emph{after} its upcoming lift-off.
Falls and terminations reinitialize the robot, so we mask a window only when its label would fall beyond a termination.
Pre-fall states whose next foothold still occurred are kept, which keeps the FPM accurate for unstable, near-failure gaits.
\subsubsection{Loss}
We train the FPM by supervised regression on the dataset $\mathcal{D}$ of recorded system states and labeled footholds, minimizing the Huber loss
\begin{equation*}
    \mathcal{L}(\theta) = \mathbb{E}_{(\mathbf{x}_t,\,\mathbf{p}_t)\sim\mathcal{D}}
    \left[ \mathcal{L}_{\delta}\!\left( f_\theta(\mathbf{x}_t) - \mathbf{p}_t \right) \right]
\end{equation*}
with threshold $\delta = 0.01$, whose bounded gradient keeps outliers near episode boundaries from dominating the fit.

\subsubsection{Network}
\label{ssec:fp_net}
The FPM is an MLP with two hidden layers with widths $[128, 128]$ followed by layer normalization and LeakyReLU$(0.1)$, with orthogonal weight initialization.
It normalizes its input with the per-dimension mean and standard deviation of $\mathbf{x}_t$ over the training split of each refit, stored in the exported model, so deployment needs no external normalization.
Table~\ref{tab:impl} lists the optimization settings.

\subsection{Foothold-Gated Height Sampling}
\label{ssec:fp_sampling}
The locomotion policy's exteroceptive input is sampled from a robot-centric elevation map $\mathcal{H}_t$~\cite{elevation_mapping, miki2022elevationcupy, pmlr-v164-rudin22a}, a 2.5D grid of $\SI{1.6}{\metre} \times \SI{1.0}{\metre}$ at \SI{0.05}{\metre} resolution ($33 \times 21 = 693$ heights) in a yaw-aligned frame attached to the base, with heights relative to the base height.
We transform each predicted foothold into the map frame, round it to the nearest cell, and gather the $k \times k$ cells around it into a patch $\mathbf{h}_t^{\mathrm{1}}, \mathbf{h}_t^{\mathrm{2}} \in \mathbb{R}^{k^2}$, with $k = 7$ and clamping at the map boundary; at \SI{0.05}{\metre} this is a \SI{0.35}{\metre} square, comparable to the \SI{0.208}{\metre} length of the Unitree G1 foot.
The two patches, each preceded by its predicted foothold, form the height scan
\begin{equation*}
    \mathbf{h}_t = \left[\ \hat{\mathbf{p}}_t^{\mathrm{1}},\ \mathbf{h}_t^{\mathrm{1}},\ \hat{\mathbf{p}}_t^{\mathrm{2}},\ \mathbf{h}_t^{\mathrm{2}}\ \right] \in \mathbb{R}^{102},
\end{equation*}
which lets the policy interpret each patch relative to its location and reduces the exteroceptive input from $693$ cells to $102$ values.
The actor observation $\mathbf{s}_t = [\mathbf{x}_t,\ \mathbf{a}_{t-1},\ \mathbf{h}_t] \in \mathbb{R}^{201}$ concatenates the system state, with angular and joint velocities scaled by $0.2$ and $0.05$, the previous action, and the height scan; the action $\mathbf{a}_t \in \mathbb{R}^{29}$ is a joint-position offset from the nominal configuration, scaled by $0.25$ and tracked by a joint-level PD controller.

To keep the policy from over-trusting the query point, we add two forms of randomization to the sampled elevation map: per-episode planar and vertical offsets in $[-0.05, 0.05]\,\si{\metre}$ displace the sampled patches uniformly while the reported coordinates stay undrifted, emulating misaligned predictions and odometry drift, and noise $\mathcal{U}(-0.1, 0.1)\,\si{\metre}$ is added to each sampled point.
The exteroceptive input is then clipped to $[-1, 1]$.
We train with an asymmetric actor-critic: the critic additionally receives the complete $693$-point map $\mathcal{H}_t$ and $\mathbf{h}_t$, so the full map informs value estimation during training, but is not needed at deployment.

\subsection{Co-Training Schedule}
\label{sec:policy}

As described in Alg.~\ref{alg:training}, we co-train the locomotion policy and the FPM in an alternating loop in which we refit each component to the latest version of the other.
The schedule runs in units of PPO iterations, each comprising $24$ environment steps for $N_{\mathrm{env}} = 8192$ parallel environments.

\subsubsection{Bootstrap}
The policy has not learned to stand or walk at the beginning of training, so no FPM is initially fit.
We instead bootstrap the locomotion policy to learn to walk on easy terrain and fix the foothold predictions directly beneath the base so that the policy observes a height scan under its own body.
We use this bootstrapping mode for the first $n_0 = 500$ iterations of RL training, at which point the policy walks adequately and its footholds carry information worth learning.

\subsubsection{FPM refit}
Every $n_\mathrm{r} = 500$ PPO iterations of locomotion policy learning, we roll out the policy and gather $N_{\mathrm{s}} = 5 \times 10^5$ transitions, which takes $\lceil N_{\mathrm{s}} / N_{\mathrm{env}} \rceil = 62$ environment steps.
For every recorded step we log the system state $\mathbf{x}_t$, the per-foot contact indicators $c_t^{\mathrm{L}}, c_t^{\mathrm{R}}$, and the world-frame base and foot poses $\mathbf{r}_t$, from which we label the future footholds $\mathbf{p}_t$ as described in Sec.~\ref{ssec:fp_target} to form the dataset $\mathcal{D}$.
We then retrain the FPM from scratch for $25$ epochs on a random $80/20$ train-validation split with shuffled mini-batches, rather than fine-tuning it, so that it cannot inherit a stale gait from earlier in training.
The refit FPM is loaded into the observation pipeline, where it gates the policy's elevation map input for the next $n_{\mathrm{r}}$ iterations before the cycle repeats.

\subsection{Training Curriculum and Implementation}
\label{ssec:terrain}

The locomotion policy is trained on procedurally generated terrains with $10$ difficulty levels and a progress curriculum similar to previous methods~\cite{pmlr-v164-rudin22a}.
The terrain suite is dominated by stepped and sparse geometry, composed of: stepping stones ($30\%$; stone width \SIrange{0.3}{0.5}{\metre}, gaps \SIrange{0.1}{0.2}{\metre}), upright and inverted pyramid stairs ($15\%$ each; step height \SIrange{0.1}{0.25}{\metre}), and boxes ($15\%$; \SI{0.45}{\metre} grid, height \SIrange{0.05}{0.3}{\metre}).
The remainder comprises upright and inverted pyramid slopes ($7.5\%$ and $5\%$, gradient up to $0.55$), random rough ground ($5\%$, noise up to \SI{0.1}{\metre}), pits ($5\%$) and flat ground ($2.5\%$).

To facilitate initial terrain traversal learning, we apply an assistive upward force to the torso equal to a fraction $\alpha(l)$ of the robot's weight, where
\begin{equation*}
    \alpha(l) = \left(1 - \beta\right)\left(1 - \frac{l}{l_{\mathrm{cut}}}\right), \quad \beta = 0.5,\ l_{\mathrm{cut}} = 4,
\end{equation*}
is clamped to $[0, 1-\beta]$ and driven by the mean terrain level $l$ across environments, the force decaying to zero at cutoff level $l_{\mathrm{cut}}$.
The robot therefore carries half its weight at level $0$ and its full weight from level $4$ onward, at which point more challenging terrains dominate.
The assistance is particularly helpful on sparse terrains given the sparse exteroceptive input, as it gives the policy room to learn to adjust a step mid-air to land on a valid foothold.

Training implementation details are presented in Table~\ref{tab:impl}, including PPO training hyperparameters and the training reward terms used.
The reward set consists of task rewards for successfully tracking the velocity command, including proper stand-still behavior, and regularization terms that minimize undesirable behavior.
Leveraging the robot's symmetry, we augment each batch with the mirrored transition instead of adding a symmetry loss~\cite{symmetry_aug}.

We use standard domain randomization parameters during training~\cite{sonic} to modify environmental and robot model parameters or apply external disturbances to add robustness to inaccurate modeling, non-ideal terrain conditions, or potential events that may occur to the robot.
In addition, we add uniform noise to all proprioceptive observations and inject occasional large spikes into the base velocity observations to emulate state-estimator faults.

\begin{table}[t]
    \vspace{5pt}
    \caption{Learning parameters for the self-supervised Foothold Prediction Model and parameters and rewards for the RL locomotion policy.}
    \label{tab:impl}\centering\footnotesize
    \setlength{\tabcolsep}{3pt}
    \vspace{-5pt}
    \begin{tabular}{@{}lr@{}}
        \toprule
        \multicolumn{2}{@{}l}{\textbf{Foothold Prediction Model (FPM)}} \\
        \midrule
        Hidden Layers / Activation                & $[128, 128]$ / LeakyReLU$(0.1)$ \\
        Input Normalization                       & z-score, train split \\
        Optimizer / Learning Rate                 & Adam / \num{5e-3} \\
        Loss                                      & Huber ($\delta = 0.01$) \\
        Batch Size / Epochs per Refit             & $1024$ / $25$ \\
        Train--Validation Split                   & $80 / 20$, random \\
        Samples per Refit $N_{\mathrm{s}}$        & $5 \times 10^5$ \\
        Bootstrap $n_0$ / Refit Interval $n_{\mathrm{r}}$ & $500$ / $500$ iterations \\
        \midrule
        \multicolumn{2}{@{}l}{\textbf{Locomotion Policy (PPO)}} \\
        \midrule
        Actor / Critic Hidden Layers              & $[512, 256, 128]$, ELU \\
        Parallel Environments $N_{\mathrm{env}}$  & $8192$ \\
        Steps per Environment per Iteration       & $24$ \\
        Learning Epochs / Mini-Batches            & $5$ / $4$ \\
        Learning Rate / Desired KL                & \num{1e-3} (adaptive) / $0.01$ \\
        Discount $\gamma$ / GAE $\lambda$         & $0.99$ / $0.95$ \\
        Clip Range / Entropy Coefficient          & $0.2$ / $0.005$ \\
        Max Gradient Norm                         & $1.0$ \\
        Initial Action Noise Std                  & $1.0$ \\
        \midrule
        \multicolumn{2}{@{}l}{\textbf{Locomotion Reward Terms}} \\
        \midrule
        $xy$ Linear Velocity Tracking (exp)       & $+1.0$ \\
        Yaw Angular Velocity Tracking (exp)       & $+1.0$ \\
        Stand Still at Nominal Pose               & $+2.5$ \\
        $z$ Linear Velocity (L2)                  & $-0.5$ \\
        $xy$ Angular Velocity (L2)                & $-0.05$ \\
        Joint Torques (L2)                        & $-10^{-5}$ \\
        Joint Accelerations (L2)                  & $-2.5\cdot10^{-8}$ \\
        Action Rate (L2)                          & $-0.01$ \\
        Undesired Contacts                        & $-1.0$ \\
        Joint Deviation Legs / Waist / Arms (L1)  & $-1.0$ / $-1.0$ / $-0.1$ \\
        Termination (Base Tilt $> \SI{0.7}{\radian}$) & $-20$ \\
        \bottomrule
    \end{tabular}
\end{table}

\section{Experiments and Results}
\label{sec:results}

\begin{figure}[t]
    \vspace{5pt}
    \centering
    \includegraphics[width=\linewidth]{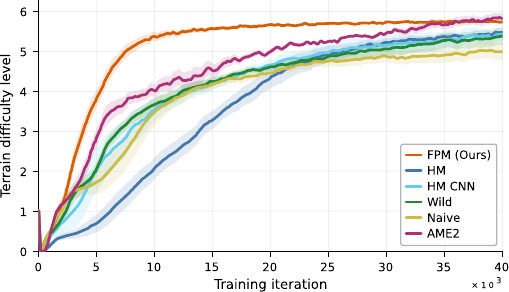}
    \vspace{-20pt}
    \caption{Mean terrain-curriculum level during training over five seeds; shaded bands show one standard deviation. FPM reaches level $5$ in $7.5$k iterations, $3.5\times$ faster than HM and $2.7\times$ faster than AME2.}
    \label{fig:learning_curve}
\end{figure}

In this section, we present simulation and hardware experiments alongside comparative results with state-of-the-art baselines and ablations regarding key choices of our method.
We leave the broader interpretation of the results to Sec.~\ref{sec:discussion}.

We compare our FPM-based locomotion policy against five alternative methods.
We evaluate five trained seeds per method over $1000$ episodes per terrain on the training terrain set and five previously unseen, out-of-distribution (OOD) terrains: undulating waves with amplitude \SIrange{0.05}{0.25}{\metre} and wavelength \SI{7.5}{\metre}, discrete rectangular obstacles \SIrange{0.5}{1.0}{\metre} tall and \SI{0.2}{\metre} wide, high-amplitude random terrain of \SIrange{0.05}{0.15}{\metre}, concentric \SI{0.35}{\metre} gaps, and sparse rectangular pallets.
All cases share the policy architecture and PPO hyperparameters of Table~\ref{tab:impl}, the same reward set, the terrain and assistive-force curricula, the same task inputs, and asymmetric critics.

The five alternative methods are as follows.
\emph{HM} provides the complete elevation map alongside the proprioceptive and task inputs.
Two alternative sampling strategies test whether a partial map around the current feet or along the commanded direction suffices: \emph{Wild} is the cylindrical height scan of Miki et al.~\cite{doi:10.1126/scirobotics.abk2822}, which queries the map around the current position of each foot, and \emph{Naive} crops a $10 \times 20$ patch of $200$ values from the map, centered at a point below the torso and shifted \SI{0.25}{\metre} per \si{\metre\per\second} of the commanded velocity, a velocity-proportional nominal placement akin to the Raibert heuristic~\cite{raibert1986legged}.
Finally, we compare against methods that train an encoder alongside the locomotion policy to extract features from the full elevation map.
\emph{HM CNN} passes the map through a CNN encoder, and \emph{AME2} is inspired by the attention-based encoder from Zhang et al.~\cite{zhang2026ame2agilegeneralizedlegged} conditioned on velocity commands.

\subsection{Learning speed}
\label{ssec:learning}

Fig.~\ref{fig:learning_curve} shows the average terrain curriculum level during training.
FPM converges the fastest, after ${\sim}15$k iterations, and reaches level $5$ after $7.5$k iterations, HM after $26.1$k, HM CNN after $25.2$k, Wild after $27.6$k, Naive after $37.9$k, and AME2 after $20.0$k.
This is a speedup of $3.4$ to $3.5\times$ over the full-map methods and $2.7\times$ over the attention encoder of AME2.
AME2 ends marginally above FPM at $40$k iterations, but at a much higher computational cost, requiring four GPUs to train and taking \SI{6.3}{\second} per policy update against \SI{0.55}{\second} for FPM, with equal rollout collection time.
FPM also shows less variance across seeds throughout training, with one seventh of the spread at $40$k iterations.

\subsection{Terrain traversal performance}
\label{ssec:locomotion}

\begin{table}[t]
    \caption{Command tracking success and fall rate for locomotion and corridor benchmarks. Stones is a training terrain, Gaps and Pallets are OOD. Best result in each column marked in \textbf{bold}, second best in \textit{italics}.}
    \label{tab:loco_comp}
    \centering\scriptsize
    \setlength{\tabcolsep}{2pt}
    \vspace{-5pt}
    \begin{tabular}{@{}l cc ccc cc cccc@{}}
        \toprule
        & \multicolumn{2}{c}{Success (\%)} & \multicolumn{3}{c}{Sparse terrain (\%)} & \multicolumn{2}{c}{Falls (\%)} & \multicolumn{4}{c}{Corridor (\%)} \\
        \cmidrule(lr){2-3}\cmidrule(lr){4-6}\cmidrule(lr){7-8}\cmidrule(l){9-12}
        Method & Train & OOD & Stones & Gaps & Pall. & Train & OOD & All & Sparse & Step. & Cont. \\
        \midrule
        FPM (ours) & \textbf{84.6} & \textit{83.1} & \textbf{80.2} & \textbf{77.7} & \textit{81.0} & \textit{3.3} & \textbf{5.2} & \textbf{89.2} & \textbf{74.2} & 85.6 & \textit{99.5} \\
        \midrule
        HM         & 81.6 & 80.8 & 69.0 & \textit{70.5} & 73.2 & 4.6 & 6.2 & 87.8 & 42.7 & \textit{93.8} & 98.7 \\
        HM CNN     & 81.9 & 78.5 & 65.7 & 58.9 & 67.9 & 5.4 & 13.3 & \textit{88.7} & 44.8 & \textbf{94.3} & \textbf{99.6} \\
        Wild       & 80.2 & 78.6 & 66.4 & 66.7 & 67.0 & 4.7 & 9.2 & 82.5 & 36.0 & 86.5 & 96.3 \\
        Naive      & 77.6 & 71.8 & 57.3 & 47.3 & 59.2 & 7.6 & 13.8 & 76.0 & 42.5 & 73.5 & 92.4 \\
        AME2       & \textit{84.3} & \textbf{83.6} & \textit{78.8} & 69.8 & \textbf{87.1} & \textbf{2.4} & \textit{5.5} & 87.3 & \textit{71.7} & 86.8 & 94.2 \\
        \bottomrule
    \end{tabular}
\end{table}

\begin{figure*}[t]
    \vspace{5pt}
    \centering
    \includegraphics[width=0.98\linewidth]{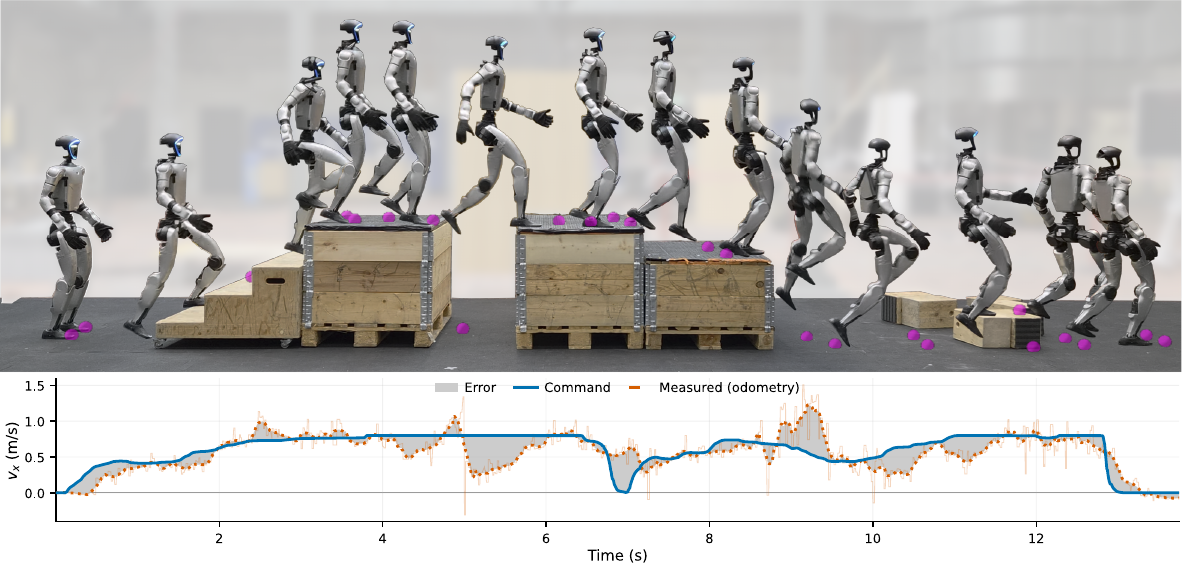}
    \vspace{-7pt}
    \caption{Sim-to-real transfer to a Unitree G1 demonstrating locomotion over challenging obstacles, including stairs and rough terrain expected within the training distribution, and a \SI{0.55}{\metre} gap and \SI{0.6}{\metre} high drop from outside the training distribution. Predicted footholds are shown in magenta for each step. The commanded x-velocity and the smoothed measured x-velocity are shown for the length of the course.}
    \label{fig:sim2real}
\end{figure*}

We compare terrain traversal performance in simulation in Table~\ref{tab:loco_comp}.
The table aggregates the results of two tests: the locomotion test, tracking a randomly sampled linear velocity command on \SI{8}{\metre} tiles, and the corridor test, tracking a forward velocity command through a narrow corridor with tall walls.
The corridor test checks how each policy handles OOD terrain outside of the robot's commanded path.
We categorize terrains as \emph{Sparse} (e.g., stepping stones), \emph{Stepped} (e.g., boxes), and \emph{Continuous} (e.g., rough).
Both tests use the terrain floors, but the corridor constrains the terrain to a \SI{1.3}{\metre} wide hall enclosed by one of three wall types: solid walls \SIrange{1.5}{2.5}{\metre} tall, a lattice of \SI{0.20}{\metre} pillars with gaps, and recessed blocks that modulate the hall's width from \SIrange{1.3}{1.5}{\metre}.

We define the success rate as the ratio of the distance traversed in the commanded direction to the distance the velocity command would have covered over the episode, so a fall counts against the full commanded distance.
Falls therefore lower the success rate, and we also report the fall rate of the locomotion test in Table~\ref{tab:loco_comp}.
A success rate of $100\%$ is not realistically achievable as it would require perfect velocity command tracking over the entire length of the course.

Our FPM policy performs best or second best in most experiments.
In the locomotion test, FPM has the highest success rate on the training terrains ($84.6\%$), the best performance on the sparse stepping stones ($80.2\%$) and gaps ($77.7\%$), and the lowest fall rate on the OOD terrains ($5.2\%$).
AME2 performs better on the OOD terrains ($83.6\%$ against $83.1\%$), including the sparse pallets ($87.1\%$), and has a lower fall rate on the training terrains ($2.4\%$).
Both policies outperform the other tested methods on aggregate performance in this test.
This demonstrates that our lightweight sampling approach matches the performance of a much more expensive attention-based approach.

The results of the corridor test are more conclusive about the benefits of the FPM sampling approach on sparse terrain.
FPM has the highest overall success rate ($89.2\%$), scoring highest on the sparse terrains ($74.2\%$) and second highest on the continuous terrains ($99.5\%$).
AME2 performs second best on the sparse terrains ($71.7\%$), and both are well ahead of all other approaches, which cover less than half the commanded distance.

\subsection{Sim-to-Real}
\label{ssec:sim2real}

We deployed our FPM-based locomotion policy in indoor and outdoor environments on a variety of terrains, as shown in the videos on the project page.
Fig.~\ref{fig:sim2real} shows the Unitree G1 traversing a course of in-distribution and OOD terrains.
The terrain consists of stairs, a \SI{0.55}{\metre} gap and a \SI{0.6}{\metre} drop outside the training distribution, and a variety of blocks.
The linear velocity command is plotted along the length of the course, showing the tracking performance relative to the obstacle.
We observe the greatest error between the commanded and measured velocity when traversing the OOD obstacles.

The policy demonstrates a variety of learned behaviors, not explicitly rewarded, that keep locomotion safe on unseen terrains.
The policy slows down before crossing the gap, which we hypothesize is caused by terrain uncertainty in the FPM-gated elevation map.
It shows similar cautious behavior before drops, slowing down and lowering its body before jumping off with one foot forward to catch itself, showing emergent behavior for terrains unseen in training.

The FPM-based locomotion policy additionally traversed a variety of outdoor terrains, including a staircase with $18$ continuous steps.
Shown in Fig.~\ref{fig:compiled_demo}, the policy also successfully traverses rough terrain in corridors, validating the simulation performance of the corridor test in Sec.~\ref{ssec:locomotion}.
The reader is encouraged to watch the videos on the project page for a full overview of the conducted hardware tests.

\subsection{Ablations}
\label{sec:ablations}
\label{ssec:horizon}
\label{ssec:fp_accuracy}

\begin{table}[t]
    \vspace{5pt}
    \caption{Ablation over prediction horizon and the addition of an exteroceptive CNN encoder. Best result in each column marked in \textbf{bold}, second best in \textit{italics}.}
    \label{tab:ablation}
    \centering\footnotesize
    \setlength{\tabcolsep}{3.5pt}
    \vspace{-5pt}
    \begin{tabular}{@{}l cc cc cccc@{}}
      \toprule
      & \multicolumn{2}{c}{Success (\%)} & \multicolumn{2}{c}{Falls (\%)} & \multicolumn{4}{c}{Corridor (\%)} \\
      \cmidrule(lr){2-3}\cmidrule(lr){4-5}\cmidrule(l){6-9}
      Method & Train & OOD & Train & OOD & All & Sparse & Step. & Cont. \\
      \midrule
      FPM & \textbf{84.6} & \textbf{83.1} & 3.3 & 5.2 & \textbf{89.2} & \textbf{74.2} & 85.6 & \textbf{99.5} \\
      FPM1 & 81.3 & 78.8 & 4.8 & 5.0 & 84.7 & 62.2 & 83.1 & 95.6 \\
      FPM3 & 79.5 & 77.3 & \textit{2.1} & 4.4 & 85.0 & 52.7 & \textit{86.3} & 96.4 \\
      FPM CNN & 82.4 & 81.1 & \textbf{1.4} & \textbf{4.0} & 85.6 & 56.2 & 85.8 & \textit{97.1} \\
      FPM1 CNN & \textit{82.7} & \textit{82.0} & 2.3 & \textit{4.2} & \textit{86.3} & \textit{62.7} & \textbf{86.7} & 95.3 \\
      FPM3 CNN & 79.3 & 78.9 & 2.2 & 6.5  & 84.8 & 53.8 & 86.2 & 95.5 \\
      \bottomrule
    \end{tabular}
\end{table}

\begin{figure}[t]
    \centering
    \includegraphics[width=\linewidth]{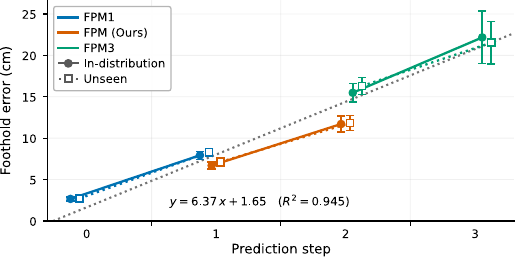}
    \vspace{-20pt}
    \caption{Foothold prediction error by prediction step of various FPM horizons for training and OOD terrains and a regression line describing the relationship between prediction step and mean error. Whiskers show one standard deviation across seeds.}
    \label{fig:pred_steps}
\end{figure}

We examine how the foothold prediction horizon affects locomotion performance.
We define two alternative horizons, denoted by the furthest step in the future that they predict, so our FPM is FPM2 under this naming.
All variants output two foothold predictions from the contact sequence but differ in the horizon.
Our FPM predicts the next two future footholds, $\mathbf{p}^{1}$ and $\mathbf{p}^{2}$, \emph{FPM1} pairs the latest foothold $\mathbf{p}^{0}$ with $\mathbf{p}^{1}$, and \emph{FPM3} outputs $\mathbf{p}^{2}$ and $\mathbf{p}^{3}$.
FPM1 tests whether the terrain around the latest and the next foothold suffices for the current policy step.
FPM3 instead explores whether a longer horizon benefits locomotion at the cost of no longer observing the next foothold.
We also test adding a CNN encoder to each FPM ablation, trained alongside the policy, to extract further features from the sampled elevation-map patches.

As shown in Table~\ref{tab:ablation}, FPM1 performs below FPM, most clearly on the sparse corridor terrains ($62.2\%$ against $74.2\%$).
The CNN encoder improves the success rate of FPM1 on both the training and OOD terrains and gives it the second best overall corridor performance ($86.3\%$).
FPM CNN has the lowest fall rate on both the training and OOD terrains ($1.4\%$ and $4.0\%$).
This suggests possible benefits of additionally encoding the sampled patches; however, FPM without an encoder still performs best overall, suggesting that the samples already form effective features for the locomotion policy.

FPM3 performs lower overall on the locomotion test.
However, FPM3 shows a lower fall rate than FPM and FPM1, and performs well on the stepped corridor terrains ($86.3\%$).
Fig.~\ref{fig:pred_steps} reports the error of each FPM horizon at each prediction step on the training and OOD terrains.
It shows that FPM3 predicts the second future foothold less accurately than FPM, and that its third-step prediction has roughly \SI{22}{\centi\metre} of error, approximately the length of the G1 foot.
Adding a CNN to FPM3 leaves success within two points of FPM3 and does not lower the OOD fall rate, likely because the further features lie too far ahead of the robot to matter for the current step.

Fig.~\ref{fig:pred_steps} also shows a roughly linear relationship between the prediction horizon and the error, of roughly \SI{6}{\centi\metre} per step.
Given the Unitree G1 foot of \SI{20.8}{\centi\metre} length and \SI{7.6}{\centi\metre} width, this suggests that the prediction is accurate to within a foot length up to the second future foothold.
The standard deviation increases with the horizon, likely due to the difficulty of predicting far ahead on rough terrain from a current motion that is only weakly related to it.

\section{Discussion and Limitations}
\label{sec:discussion}

Our co-trained, state-based sampling shrinks the perception learning problem: rather than finding the relevant cells in the full map, the policy learns how to use two small patches chosen for it.
FPM learns faster than the tested full-map methods (Sec.~\ref{ssec:learning}).
Wild and Naive, which sample as many heights as FPM or more, but around the current feet or along the commanded direction, perform worse than HM (Table~\ref{tab:loco_comp}), suggesting that the benefit comes from placing the patches at the predicted footholds, not from the smaller observation.
Despite the limited exteroceptive observations, the FPM-based policy performs best on most tested sparse terrains.
On hardware, it slows down and adjusts its body and steps before crossing large gaps or stepping off high drops (Sec.~\ref{ssec:sim2real}), demonstrating emergent cautious behavior.
We conjecture that because the observations depend on the locomotion policy's actions, this may encourage it to probe the terrain before committing while still tracking the velocity command.

AME2 is the only alternative that matches FPM on the locomotion test, at a much higher training cost of \SI{6.3}{\second} per policy update on four GPUs compared to \SI{0.55}{\second} on a single GPU for FPM.
Its weaker corridor results (Sec.~\ref{ssec:locomotion}) suggest that the attention mechanism can be distracted by obstacles it has never seen, whereas the FPM never shows the policy the wall outside the commanded direction.
The corridor test also marks the limits of our approach.
Full-map observation methods lead FPM on the stepped terrains, and all methods perform similarly on continuous terrain, where a dense scan suffices.
The FPM discards all terrain the robot does not intend to step on, and obstacles approaching from the side would enter the patches late or not at all, which our corridor walls, parallel to the commanded path, do not test.

Finally, our effective prediction horizon is limited to predicting the second future foothold.
The FPM3 ablation (Fig.~\ref{fig:pred_steps}) shows that the step predictions become less accurate with greater variance as the horizon increases, also degrading the locomotion policy's performance (Table~\ref{tab:ablation}).

\section{Conclusion}
\label{sec:conclusion}

In this paper, we presented a perceptive locomotion framework for humanoids in which a lightweight foothold prediction model, trained alongside the locomotion policy, predicts the next two footholds from proprioception and the velocity command, and we sample the elevation map only around them.
The policy learns several times faster than policies that observe the complete map, matches an attention-based encoder at a fraction of its training cost, performs best on most sparse terrains, and transfers zero-shot to a Unitree G1.

In the future this work may be extended to an active depth-based perception system that does not rely on an elevation map, but instead directly observes the future predicted footholds.
This approach may also extend to manipulative locomotion, using predicted hand and foot contacts to guide the policy to observe points for arm and leg support.

\section*{ACKNOWLEDGMENT}
The authors acknowledge the use of generative AI to assist with code generation and language refinement.
We thank Tianxu An, Pwint Thu Aung, Dario Hug, Shengzhi Wang, and Fan Yang for their assistance with testing and filming, Victor Klemm for discussions that inspired the assistive force curriculum, and René Zurbrügg for his help in preparing the visuals for this work.

\bibliographystyle{IEEEtran}
\bibliography{references}

\begin{thebibliography}{10}
\providecommand{\url}[1]{#1}
\csname url@samestyle\endcsname
\providecommand{\newblock}{\relax}
\providecommand{\bibinfo}[2]{#2}
\providecommand{\BIBentrySTDinterwordspacing}{\spaceskip=0pt\relax}
\providecommand{\BIBentryALTinterwordstretchfactor}{4}
\providecommand{\BIBentryALTinterwordspacing}{\spaceskip=\fontdimen2\font plus
\BIBentryALTinterwordstretchfactor\fontdimen3\font minus \fontdimen4\font\relax}
\providecommand{\BIBforeignlanguage}[2]{{%
\expandafter\ifx\csname l@#1\endcsname\relax
\typeout{** WARNING: IEEEtran.bst: No hyphenation pattern has been}%
\typeout{** loaded for the language `#1'. Using the pattern for}%
\typeout{** the default language instead.}%
\else
\language=\csname l@#1\endcsname
\fi
#2}}
\providecommand{\BIBdecl}{\relax}
\BIBdecl

\bibitem{pmlr-v164-rudin22a}
N.~Rudin \emph{et~al.}, ``Learning to walk in minutes using massively parallel deep reinforcement learning,'' in \emph{Proc. Conf. Robot Learn. (CoRL)}, 2022, pp. 91--100.

\bibitem{doi:10.1126/scirobotics.abk2822}
T.~Miki \emph{et~al.}, ``Learning robust perceptive locomotion for quadrupedal robots in the wild,'' \emph{Sci. Robot.}, vol.~7, no.~62, p. eabk2822, 2022.

\bibitem{elevation_mapping}
P.~Fankhauser \emph{et~al.}, ``Robot-centric elevation mapping with uncertainty estimates,'' in \emph{Mobile Service Robotics}.\hskip 1em plus 0.5em minus 0.4em\relax World Scientific, 2014, pp. 433--440.

\bibitem{FootPlacementLookControl}
J.~S. Matthis, J.~L. Yates, and M.~M. Hayhoe, ``Gaze and the control of foot placement when walking in natural terrain,'' \emph{Curr. Biol.}, vol.~28, no.~8, pp. 1224--1233.e5, 2018.

\bibitem{hoeller2024anymalparkour}
D.~Hoeller \emph{et~al.}, ``{ANYmal} parkour: Learning agile navigation for quadrupedal robots,'' \emph{Sci. Robot.}, vol.~9, no.~88, p. eadi7566, 2024.

\bibitem{cheng2023parkour}
X.~Cheng \emph{et~al.}, ``Extreme parkour with legged robots,'' in \emph{Proc. IEEE Int. Conf. Robot. Autom. (ICRA)}, 2024, pp. 11\,443--11\,450.

\bibitem{long2024perceptiveinternalmodel}
J.~Long \emph{et~al.}, ``Learning humanoid locomotion with perceptive internal model,'' in \emph{Proc. IEEE Int. Conf. Robot. Autom. (ICRA)}, 2025, pp. 9997--10\,003.

\bibitem{ben2025gallant}
Q.~Ben \emph{et~al.}, ``Gallant: Voxel grid-based humanoid locomotion and local-navigation across {3-D} constrained terrains,'' in \emph{Proc. IEEE/CVF Conf. Comput. Vis. Pattern Recognit. (CVPR)}, 2026, pp. 28\,086--28\,095.

\bibitem{doi:10.1126/scirobotics.adv3604}
J.~He \emph{et~al.}, ``Attention-based map encoding for learning generalized legged locomotion,'' \emph{Sci. Robot.}, vol.~10, no. 105, p. eadv3604, 2025.

\bibitem{zhang2026ame2agilegeneralizedlegged}
C.~Zhang \emph{et~al.}, ``{AME-2}: Agile and generalized legged locomotion via attention-based neural map encoding,'' arXiv:2601.08485, 2026.

\bibitem{fu2026globallocalattentiondecompositionterrain}
S.~Fu \emph{et~al.}, ``Global-local attention decomposition for terrain encoding in humanoid perceptive locomotion,'' arXiv:2606.00637, 2026.

\bibitem{li2026taga}
P.~Li \emph{et~al.}, ``{TAGA}: Terrain-aware active gaze learning for generalizable agile humanoid locomotion,'' arXiv:2606.05880, 2026.

\bibitem{zhuang2024humanoid}
Z.~Zhuang, S.~Yao, and H.~Zhao, ``Humanoid parkour learning,'' in \emph{Proc. Conf. Robot Learn. (CoRL)}, 2024.

\bibitem{yu2026ssr}
R.~Yu \emph{et~al.}, ``{SSR}: Scaling surefooted and symmetric humanoid traversal to the open world,'' arXiv:2605.30770, 2026.

\bibitem{6907083}
D.~Kanoulas and M.~Vona, ``Bio-inspired rough terrain contact patch perception,'' in \emph{Proc. IEEE Int. Conf. Robot. Autom. (ICRA)}, 2014, pp. 1719--1724.

\bibitem{KANOULAS201913}
D.~Kanoulas, N.~G. Tsagarakis, and M.~Vona, ``Curved patch mapping and tracking for irregular terrain modeling: Application to bipedal robot foot placement,'' \emph{Robot. Auton. Syst.}, vol. 119, pp. 13--30, 2019.

\bibitem{10.1109/TRO.2022.3222958}
S.~Fahmi \emph{et~al.}, ``{ViTAL}: Vision-based terrain-aware locomotion for legged robots,'' \emph{IEEE Trans. Robot.}, vol.~39, no.~2, pp. 885--904, 2023.

\bibitem{WANG2024104742}
K.~Wang \emph{et~al.}, ``When and where to step: Terrain-aware real-time footstep location and timing optimization for bipedal robots,'' \emph{Robot. Auton. Syst.}, vol. 179, p. 104742, 2024.

\bibitem{magana2019footholdcnn}
O.~A. {Villarreal Maga{\~n}a} \emph{et~al.}, ``Fast and continuous foothold adaptation for dynamic locomotion through {CNNs},'' \emph{IEEE Robot. Autom. Lett.}, vol.~4, no.~2, pp. 2140--2147, 2019.

\bibitem{belter2019singleshot}
D.~Belter \emph{et~al.}, ``Single-shot foothold selection and constraint evaluation for quadruped locomotion,'' in \emph{Proc. IEEE Int. Conf. Robot. Autom. (ICRA)}, 2019, pp. 7441--7447.

\bibitem{xie2020allsteps}
Z.~Xie \emph{et~al.}, ``{ALLSTEPS}: Curriculum-driven learning of stepping stone skills,'' \emph{Comput. Graph. Forum}, vol.~39, no.~8, pp. 213--224, 2020.

\bibitem{wang2025beamdojo}
H.~Wang \emph{et~al.}, ``{BeamDojo}: Learning agile humanoid locomotion on sparse footholds,'' in \emph{Proc. Robot.: Sci. Syst. ({RSS})}, 2025.

\bibitem{jenelten2024dtc}
F.~Jenelten \emph{et~al.}, ``{DTC}: Deep tracking control,'' \emph{Sci. Robot.}, vol.~9, no.~86, p. eadh5401, 2024.

\bibitem{montenegro2026mindstepsgenerallearning}
A.~Montenegro \emph{et~al.}, ``Mind your steps: A general learning framework for accurate humanoid foothold tracking,'' in \emph{Proc. Robot.: Sci. Syst. ({RSS})}, 2026.

\bibitem{dai2026walkplancphysicsguidedrl}
M.~Dai \emph{et~al.}, ``Walk the {PLANC}: Physics-guided {RL} for agile humanoid locomotion on constrained footholds,'' arXiv:2601.06286, 2026.

\bibitem{2019deepgait}
V.~Tsounis \emph{et~al.}, ``{DeepGait}: Planning and control of quadrupedal gaits using deep reinforcement learning,'' \emph{IEEE Robot. Autom. Lett.}, vol.~5, no.~2, pp. 3699--3706, 2020.

\bibitem{PPO}
J.~Schulman \emph{et~al.}, ``Proximal policy optimization algorithms,'' arXiv:1707.06347, 2017.

\bibitem{nvidia2025isaaclabgpuacceleratedsimulation}
NVIDIA \emph{et~al.}, ``{Isaac Lab}: A {GPU}-accelerated simulation framework for multi-modal robot learning,'' arXiv:2511.04831, 2025.

\bibitem{miki2022elevationcupy}
T.~Miki \emph{et~al.}, ``Elevation mapping for locomotion and navigation using {GPU},'' in \emph{Proc. IEEE/RSJ Int. Conf. Intell. Robots Syst. (IROS)}.\hskip 1em plus 0.5em minus 0.4em\relax IEEE, 2022, pp. 2273--2280.

\bibitem{symmetry_aug}
M.~Mittal \emph{et~al.}, ``Symmetry considerations for learning task symmetric robot policies,'' in \emph{Proc. IEEE Int. Conf. Robot. Autom. (ICRA)}.\hskip 1em plus 0.5em minus 0.4em\relax IEEE, 2024, pp. 7433--7439.

\bibitem{sonic}
Z.~Luo \emph{et~al.}, ``{SONIC}: Supersizing motion tracking for natural humanoid whole-body control,'' \emph{Sci. Robot.}, vol.~11, no. 117, p. eaed4592, 2026.

\bibitem{raibert1986legged}
M.~H. Raibert, \emph{Legged Robots That Balance}.\hskip 1em plus 0.5em minus 0.4em\relax Cambridge, MA: MIT Press, 1986.

\end{thebibliography}
\end{document}